\documentclass[11pt,a4paper]{article}

\usepackage[utf8]{inputenc}
\usepackage[T1]{fontenc}

\usepackage{newtxtext,newtxmath}

\usepackage[
    top=1in,
    bottom=1in,
    left=1in,
    right=1in
]{geometry}

\usepackage{authblk}

\usepackage{amsmath}
\usepackage{mathtools}

\usepackage{graphicx}
\usepackage{subcaption}

\usepackage{tikz}
\usetikzlibrary{positioning,arrows.meta,calc,fit}

\usepackage{booktabs}
\usepackage{multirow}
\usepackage{array}

\usepackage{siunitx}

\usepackage{enumitem}

\usepackage{microtype}

\usepackage{xcolor}

\usepackage[
    colorlinks=true,
    linkcolor=blue,
    citecolor=blue,
    urlcolor=blue
]{hyperref}

\usepackage[
    nameinlink,
    capitalise
]{cleveref}

\usepackage[
    backend=biber,
    style=ieee,
    sorting=none
]{biblatex}

\makeatletter
\g@addto@macro\normalsize{%
    \setlength{\abovedisplayskip}{8pt plus 2pt minus 2pt}%
    \setlength{\belowdisplayskip}{8pt plus 2pt minus 2pt}%
    \setlength{\abovedisplayshortskip}{8pt plus 2pt minus 2pt}%
    \setlength{\belowdisplayshortskip}{8pt plus 2pt minus 2pt}%
}
\makeatother

\graphicspath{{figures/}}

\title{
    \Large
    \textbf{The Effect of Missingness-Pattern Mismatch on Method Selection for Time-Series Classification: A Controlled Empirical Study}
}

\author[1,$\dagger$]{Ruiqi Zhao}
\author[2]{Zishun Yuan}
\author[3]{Zhentao Wang}
\author[4]{Jiahao Quan}
\author[5]{Kangzheng Li}
\author[6]{Jianfan Deng}

\DeclareRobustCommand{\graydagger}{{$\dagger$}}
\affil[{}\graydagger]{{Corresponding Author}}

\affil[1]{
The University of Tokyo,
Tokyo, Japan\\
\href{mailto:richard-zhao@g.ecc.u-tokyo.ac.jp}
{\texttt{richard-zhao@g.ecc.u-tokyo.ac.jp}}
}

\affil[2]{
Stony Brook University, SUNY Korea,
Incheon, South Korea\\
\href{mailto:yuan.zishun@stonybrook.edu}
{\texttt{yuan.zishun@stonybrook.edu}}
}

\affil[3]{
Fuyao University of Science and Technology, Fujian, China\\
\href{mailto:zane.wang@stu.fyust.edu.cn}
{\texttt{zane.wang@stu.fyust.edu.cn}}
}

\affil[4]{
Renmin University of China,
Beijing, China\\
\href{mailto:qjh314@ruc.edu.cn}
{\texttt{qjh314@ruc.edu.cn}}
}

\affil[5]{
Tsinghua University,
Beijing, China\\
\href{mailto:likz26@mails.tsinghua.edu.cn}
{\texttt{likz26@mails.tsinghua.edu.cn}}
}

\affil[6]{
Anhui Science and Technology University,
Anhui, China\\
\href{mailto:miku0831@ahstu.edu.cn}
{\texttt{miku0831@ahstu.edu.cn}}
}

\date{}

\begin{document}

\maketitle


\begin{abstract}
Classifiers for time-series classification are commonly selected on validation data, but the temporal pattern of missing observations at deployment may differ from the pattern seen during validation. We examine whether such a mismatch affects validation-based classifier selection. In a controlled $2 \times 2$ design, validation and test sets of 64 univariate UCR datasets were masked with either random point missingness or circular block missingness at six rates from 5\% to 30\%, imputed by linear interpolation, and used to select among three prespecified candidates: 1NN-DTW, MiniRocket with a Ridge classifier, and a statistical-feature Random Forest. Training data remained complete, and selections made under matched and mismatched validation patterns were compared on the same masked test sets. Mismatched validation reduced the test balanced accuracy of the selected classifier by 1.14 percentage points on average (95\% CI 0.79 to 1.51), with losses on 49 of the 64 datasets. The loss was negligible at 5\% missingness and increased to 2.46 percentage points at 30\%. It was concentrated in point-masked deployment (1.84 percentage points), where block-masked validation shifted selection away from the usually best candidate, while the effect for block-masked deployment was small and not significant. Mismatch changed the selected classifier in 35.5\% of paired comparisons, but a changed selection did not always reduce performance. A supplementary analysis with non-wrapping linear blocks reproduced these findings with a larger effect (1.67 percentage points). Matching the missingness pattern of validation data to the expected deployment pattern is therefore a simple safeguard for method selection, particularly at higher missingness rates.
\end{abstract}


\section{Introduction}
\label{sec:introduction}

Time-series classification (TSC) assigns labels to sequences whose temporal order may be important for distinguishing classes. Public data archives and multi-dataset benchmarks have made comparisons among TSC methods more systematic and reproducible~\cite{dau2019ucr,bagnall2017,middlehurst2024bakeoff}. Yet benchmark conclusions depend on the evaluation protocol: choices about data splitting, preprocessing, and evaluation design can affect estimated performance and, in turn, which method is selected for a designated task~\cite{bagnall2017}. Whether a choice made under one validation condition remains appropriate under a different test condition is therefore an empirical question.

Missing observations are one source of variation between validation and test conditions. Measurements may be absent either as individual observations at \textbf{randomly selected time points} (\textbf{point missingness}) or as contiguous runs of missing observations (\textbf{block missingness}). These two missingness patterns can remove the same number of observations while producing different temporal structures in the observed data. Time-series imputation benchmarks have evaluated methods under different missingness rates, mechanisms, and patterns~\cite{kazijevs2023deep,poette2026benchmarking}, while studies of incomplete TSC have examined the consequences of missing observations for classification performance~\cite{ma2022adversarial}. Missingness rate alone does not fully describe the conditions under which a method is evaluated.

Missingness conditions may also differ between model development and deployment. Previous work has examined prediction when missingness mechanisms differ between source and target data~\cite{zhou2023domain,rockenschaub2024robust}. Separate benchmarking studies have evaluated time-series imputation under different missingness rates and temporal missingness scenarios, including isolated and contiguous gaps, and have shown that method performance can depend on the missingness scenario~\cite{poette2026benchmarking}. Other recent work has formulated pipeline selection for incomplete data as an algorithm-selection problem, motivated by differences in method performance across missingness profiles~\cite{li2026prediction}. Taken together, these studies show that missingness can affect both predictive performance and method choice. However, to our knowledge, the consequences of validation--test missingness-pattern mismatch for classifier selection in TSC remain insufficiently characterized. We refer to a validation condition as \emph{matched} when its missingness-pattern type is the same as that of the target test condition, and as \emph{mismatched} when the validation and test pattern types differ. In particular, it remains unclear whether selecting among classifiers under one missingness pattern leads to a different classifier-selection decision when the selected method is evaluated under another.

The distinction is between a classifier's performance and the decision used to select a classifier. A classifier may score highest on validation sequences with point missingness, while another performs better on test sequences with block missingness. The central question is whether the validation protocol leads to a worse selection decision when classifier performance changes under missingness. A matched validation protocol can still produce an imperfect choice when the available validation data are limited, even when the validation and test patterns agree~\cite{cawley2010overfitting}. Comparing matched and mismatched selection strategies on the same target test condition therefore provides a direct reference for quantifying the performance difference associated with mismatch.

We ask: \textbf{At a fixed missingness rate, how does a mismatch between validation and test missingness patterns affect the selected classifier and its test performance?} We compare point missingness with circular block missingness in a controlled $2 \times 2$ validation--test design. For each target test condition, candidate classifiers are selected using either a matched or a mismatched validation pattern and then evaluated on the same test instances under the same target missingness masks. The circularity applies only to the masking construction, allowing a contiguous block to wrap around sequence boundaries, while the underlying time series are treated without any periodicity assumption. The primary outcomes are the paired differences in test balanced accuracy between classifiers selected under matched and mismatched validation conditions, reported separately for point and block test conditions.

We examine this question across public, univariate, equal-length TSC datasets using three prespecified candidate classification methods (1NN-DTW~\cite{bagnall2014nn,bagnall2017}, MiniRocket with a Ridge classifier~\cite{dempster2021minirocket}, and a statistical-feature-based Random Forest~\cite{breiman2001random}) and a fixed linear-interpolation procedure. We also assess how often the selected method changes and quantify \textbf{selection regret}, defined as the difference between the test balanced accuracy of the retrospectively best-performing candidate under the target test condition and that of the classifier selected by validation. We further examine how these quantities vary across tasks and missingness rates. Mask generation, preprocessing, selection, and evaluation procedures are specified and documented to permit replication. Our aim is to assess the reliability of validation-based method selection across target test conditions with different missingness patterns and across a diverse set of candidate classifiers.

\section{Related Work}
\label{sec:related}

\subsection{Time-Series Classification and Method Comparison}

Time-series classification has been studied through distance-based, feature-based, and transformation-based methods. Public benchmark archives have made it possible to compare these approaches across classification-tasks collections under shared evaluation protocols. The UCR Time Series Classification Archive provides a widely-used benchmark for this purpose, while subsequent studies have compared nearest-neighbour methods and a broader set of TSC algorithms across multiple datasets~\cite{dau2019ucr,bagnall2017,middlehurst2024bakeoff}. These benchmark studies also show that conclusions about relative classifier performance depend on the evaluation procedures and the datasets included in the comparison~\cite{bagnall2017}.

Nearest-neighbour classification remains a useful reference point in TSC, particularly when distance measures capture relevant temporal structure. Bagnall and Lines evaluated nearest-neighbour classifiers with several time-series distance measures and found substantial variation in performance across datasets~\cite{bagnall2014nn}. Our distance-based candidate is one-nearest-neighbour classification with dynamic time warping (1NN-DTW), which remains the standard benchmark against which new TSC algorithms are compared~\cite{middlehurst2024bakeoff}. Other methods construct alternative representations before classification. MiniRocket, for example, uses convolutional transformations followed by a linear classifier and provides a computationally efficient TSC method~\cite{dempster2021minirocket}. In our study, MiniRocket is paired with a Ridge classifier. Feature-based TSC represents time series using derived descriptive properties, or features, and learns a classifier in the resulting feature space~\cite{fulcher2014feature}. We combine statistical features with a Random Forest classifier~\cite{breiman2001random}, giving a representation that differs from both direct distance comparison and convolutional transformation.

These methods form the prespecified candidate set in the present study. Their purpose is to examine whether the validation-based selection outcome and the resulting test performance can change when the temporal pattern of missing observations changes.

\subsection{Missing Data in Time-Series Analysis}

Missing observations are common in time-series data, and their effects can depend on both the amount and the structure of missingness. Imputation benchmarks have varied missingness rates and the statistical missingness mechanisms formalized by Rubin~\cite{rubin1976inference,little2019statistical}, namely missing completely at random (MCAR), missing at random (MAR), and missing not at random (MNAR)~\cite{kazijevs2023deep}; these mechanisms are conceptually distinct from the temporal arrangement of missing positions. More recent benchmarking has explicitly contrasted randomly missing observations with structured temporal gaps, such as those caused by sensor disconnections and workflow-driven interruptions~\cite{poette2026benchmarking}, while dedicated time-series recovery benchmarks have examined contiguous missing blocks~\cite{khayati2020mind}. Together, these studies indicate that missingness cannot always be characterized adequately by its overall proportion.

The temporal arrangement of missing observations can change the information available to a downstream classifier. Individual missing points and contiguous gaps can remove the same number of observations while altering distinct parts of the local temporal structure. Consequently, two differently masked versions of the same series can present different inputs to a classifier after imputation, even when the nominal missingness rate is the same.

Incomplete time-series classification has also been studied directly. Ma et al.\ proposed a recurrent approach for classification with incomplete time series and evaluated prediction under missing observations as a part of the classification problem rather than as a preprocessing detail alone~\cite{ma2022adversarial}. More recently, Kim et al.\ proposed a probabilistic framework for classifying multivariate time series with missing values that combines a deep generative imputation model with a classifier and accounts for predictive uncertainty arising from multiple plausible imputations~\cite{kim2023probabilistic}. By contrast, in our study the imputation procedure is fixed (linear imputation). We examine how changes in the missingness pattern affect classifier selection under that procedure, which allows us to avoid comparing imputation methods.

\subsection{Missingness Shift Across Development and Test Conditions}

A related body of work considers cases in which the missingness process changes between datasets or between development and deployment conditions. Zhou et al.\ studied domain adaptation under missingness shift, where the relationship between observed and missing values can differ between source and target domains~\cite{zhou2023domain}. Rockenschaub et al.\ examined prediction under changing missingness conditions and showed that predictive behaviour can vary when the missingness distribution changes between development and deployment settings~\cite{rockenschaub2024robust}.

These studies motivate attention to changes in missingness conditions across stages of a modelling pipeline. The problem considered here is more specific. For each dataset and random seed, the official UCR training set is split into a complete fitting subset and a held-out validation subset. The candidate classifiers are trained on the complete fitting subset, while missingness is introduced only into the held-out validation subset and the official test set. A classifier is selected using the masked validation data and then evaluated on a fixed target test condition. Thus, the study concerns how the validation condition used for method selection relates to the target test condition, rather than a change in the missingness of the training data.

Related work has also considered the choice of prediction pipelines under different missingness profiles. Li et al.\ used missingness characteristics to support pipeline selection in clinical prediction~\cite{li2026prediction}. This work is relevant to the idea that missing-data conditions can affect method choice. Our study instead uses a controlled TSC setting in which the candidate classifiers and imputation procedure are fixed while the temporal missingness pattern observed during validation is varied.

A recent public software benchmark and reproducibility repository has also examined model-selection transport under broader missing-data distribution shifts, including candidate-ordering stability, winner identity, and oracle-relative selection regret~\cite{kovari2026missingnessshiftbench}. Its setting is not specific to time-series classification or temporal point-versus-block missingness.

\subsection{Positioning of the Present Study}

Prior work has established that TSC methods can have different relative performance across evaluation settings~\cite{dau2019ucr,bagnall2017,bagnall2014nn,dempster2021minirocket}, that missingness conditions and their temporal structure can matter for downstream modelling~\cite{kazijevs2023deep,poette2026benchmarking,ma2022adversarial}, and that changes in missingness conditions can alter predictive behaviour~\cite{zhou2023domain,rockenschaub2024robust,li2026prediction}. The question addressed in this study is much narrower: at a fixed missingness rate and target test condition, does changing the validation missingness pattern alter the classifier selected by the validation procedure, and how does that change affect test balanced accuracy.

Prediction performance and method selection are distinct questions. Changing the validation pattern may leave test balanced accuracy unchanged if the selected classifiers perform equally well on the same masked test set, so a change in classifier identity does not by itself imply a difference in test performance.

For each fixed target test condition, we compare a matched and mismatched validation condition. When the target test pattern is point missingness, point-missing validation and block-missing validation respectively produce the matched and mismatched selections. When the target test pattern is block missingness, block-missing validation and point-missing validation form the corresponding matched and mismatched pair. The two selections are assessed on the same masked test set so that the primary comparison is the paired difference in test balanced accuracy associated with changing the validation pattern under the specified selection procedure.

The study uses point missingness and circular block missingness (so that every time point has the same probability of being masked). Circular block masking selects a contiguous block under circular indexing, so a block may extend from the end of a series to its beginning. The circular indexing applies only to mask construction. The underlying time series are not treated as periodic ones; the interpolation procedure operates on the original linear time axis. The formal masking procedure is given in \cref{sec:missingness-generation}.

Selection regret provides a complementary measure of the selected method's test performance relative to the test oracle. It is defined as the difference between the highest balanced accuracy achieved by any candidate classifier on the same target test set and the balanced accuracy of the classifier selected from validation. The test oracle is determined retrospectively from test labels and serves only as an evaluation reference; it does not participate in classifier selection.

The primary comparison is the paired difference in test balanced accuracy between matched and mismatched validation-based selections. We record whether the selected classifier changes and report selection regret for each validation-based choice as well. These quantities describe different aspects of the selection process: The primary comparison measures the dissimilarity in test performance between the two validation conditions, classifier identity describes the resulting selection decision, and regret measures the gap between that decision and the highest-performing candidate classifier on the target test set.

\section{Methodology}
\label{sec:methodology}

\subsection{Research Question}
\label{sec:research-question}

This study examines whether a mismatch between validation and test missingness patterns affects validation-based classifier selection in time-series classification. At a fixed nominal missingness rate, we compare selections made under matched and mismatched validation patterns while holding the target test condition fixed.

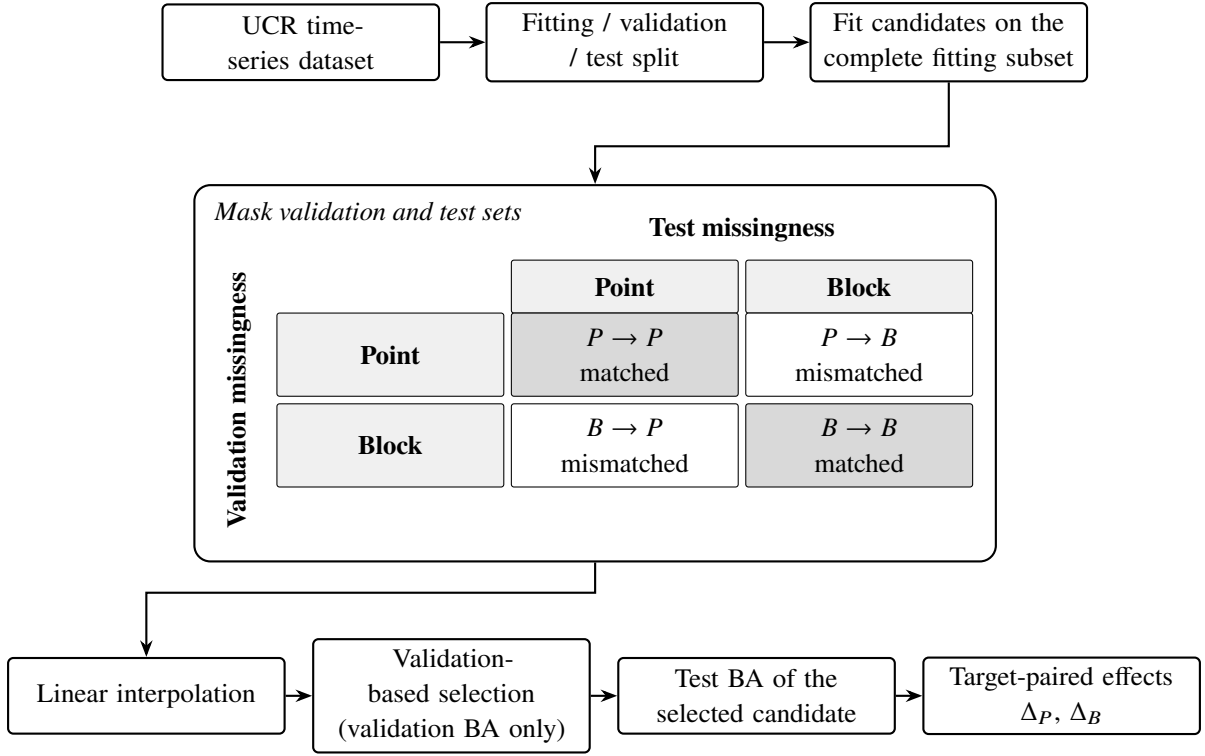
\begin{figure}[t]
\centering
\begin{tikzpicture}[
    font=\small,
    box/.style={draw, thick, rounded corners=2pt, align=center,
                minimum height=10mm, text width=34mm,
                execute at begin node={\hyphenpenalty=10000}},
    cell/.style={draw, rounded corners=1pt, align=center,
                 minimum width=30mm, minimum height=11mm},
    head/.style={cell, fill=gray!12, font=\small\bfseries, minimum height=7mm},
    rowhead/.style={cell, fill=gray!12, font=\small\bfseries},
    matched/.style={cell, fill=gray!30},
    mismatched/.style={cell, fill=white},
    arr/.style={-{Stealth[length=2.2mm]}, thick},
    node distance=6mm and 6mm,
]
\node[box] (split) {Fitting / validation / test split};
\node[box, left=of split] (data) {UCR time-series dataset};
\node[box, right=of split] (fit) {Fit candidates on the\\complete fitting subset};
\draw[arr] (data) -- (split);
\draw[arr] (split) -- (fit);

\begin{scope}[shift={($(split.south)+(0,-27mm)$)}]
    \node[head] (hp) at (0,0)        {Point};
    \node[head] (hb) at (31mm,0)     {Block};
    \node[rowhead] (rp) at (-31mm,-9mm) {Point};
    \node[rowhead] (rb) at (-31mm,-21mm) {Block};
    \node[matched]    (pp) at (0,-9mm)     {$P \to P$\\matched};
    \node[mismatched] (pb) at (31mm,-9mm)  {$P \to B$\\mismatched};
    \node[mismatched] (bp) at (0,-21mm)    {$B \to P$\\mismatched};
    \node[matched]    (bb) at (31mm,-21mm) {$B \to B$\\matched};
    \node[font=\small\bfseries, above=1mm of $(hp.north)!0.5!(hb.north)$] (tl) {Test missingness};
    \node[font=\small\bfseries, rotate=90, left=2mm of $(rp.west)!0.5!(rb.west)$, anchor=south]
        (vl) {Validation missingness};
\end{scope}
\node[draw, thick, rounded corners=6pt, inner sep=3mm, fit=(tl) (vl) (pp) (bb) (hb) (rb)] (frame) {};
\node[font=\small\itshape, anchor=north west] at ($(frame.north west)+(1.5mm,-1mm)$)
    {Mask validation and test sets};
\draw[arr] (fit.south) |- ($(frame.north)+(0,5mm)$) -- (frame.north);

\node[box, below=10mm of frame.south, xshift=-19mm] (sel) {Validation-based selection\\(validation BA only)};
\node[box, left=3.5mm of sel]  (imp)  {Linear interpolation};
\node[box, right=3.5mm of sel] (eval) {Test BA of the\\selected candidate};
\node[box, right=3.5mm of eval] (eff) {Target-paired effects\\$\Delta_P$, $\Delta_B$};
\draw[arr] (frame.south) -- ++(0,-4mm) -| (imp.north);
\draw[arr] (imp) -- (sel);
\draw[arr] (sel) -- (eval);
\draw[arr] (eval) -- (eff);
\end{tikzpicture}
\caption{Experimental design. Candidates are fitted on the complete (unmasked) fitting subset. Validation and test sets are masked with point ($P$) or circular block ($B$) missingness and imputed by linear interpolation; the candidate with the highest validation balanced accuracy (BA) is then evaluated on the test set. Grey cells are matched and white cells mismatched conditions. $\Delta_P$ and $\Delta_B$ are the target-paired effects defined in \cref{sec:mismatch-effect}.}
\label{fig:design}
\end{figure}

The primary experiment uses a controlled $2 \times 2$ design with point missingness ($P$) and circular block missingness ($B$). The four conditions are $PP$, $PB$, $BP$, and $BB$, where the first letter denotes the validation pattern and the second the test pattern. A condition is \emph{matched} when the two pattern types agree and \emph{mismatched} otherwise; matching does not require identical validation and test mask realizations. \Cref{fig:design} summarizes the design.

The primary outcome is the paired difference in test balanced accuracy between the selected candidates. Candidate identity is also recorded, because a change in the selected method need not change test performance when the candidates achieve equal scores on the same target test set.

\subsection{Datasets and Data Splitting}
\label{sec:datasets}

The primary experiment uses 64 univariate, equal-length, originally complete datasets from the 85-dataset UCR2015 archive~\cite{dau2019ucr}, accessed through the aeon toolkit~\cite{aeon24jmlr} (version 1.6.0). The final dataset list was fixed before formal result analysis and is recorded in the study protocol and configuration.

An initial resource-based screening omitted 21 datasets whose combinations of sample size and series length made 1NN-DTW impractical. Loading and stratified-splitting checks subsequently identified four unusable datasets in the initial selection, and four of the initially omitted datasets were included as replacements. The final exclusions therefore comprise 17 datasets omitted for computational reasons and four that could not be loaded or split under the specified procedure. Neither selection nor replacement used classification performance or missingness outcomes.

Each instance is a labelled sequence $x = (x_1, \ldots, x_T)$. The length $T$ is constant within each dataset but may differ across datasets. For each dataset and seed, the official UCR training set was divided into a complete fitting subset and a held-out validation subset using a stratified split with a target validation fraction of 25\%; the realized fraction may differ slightly because instance counts are integers. Five seeds, $\{1, 2, 3, 4, 5\}$, were used.

The official UCR test set was retained as the target test set. Within a dataset and seed, the same train--validation split was reused across all rates and patterns. Fitting data remained complete; synthetic missingness was introduced only into validation and test data.

Descriptive outputs from the first 27 and 46 completed datasets were inspected during the primary run. These were ordered prefixes of the fixed dataset list, not separately selected samples, and were not used as the final inferential sample.

\subsection{Missingness Generation}
\label{sec:missingness-generation}

For a series of length $T$ and nominal missingness rate $r$, the number of masked observations was
\begin{equation}
    k = \lfloor rT \rfloor .
\end{equation}
The same integer $k$ was used for all patterns at a given length and rate, and both $r$ and the realized rate $k/T$ were recorded.

In mathematical notation, the observation mask is
\begin{equation}
    m_t =
    \begin{cases}
        1, & \text{observed},\\
        0, & \text{missing}.
    \end{cases}
\end{equation}
Positions with $m_t = 0$ were replaced by missing values before imputation. The implementation stores the complementary Boolean indicator, with \texttt{True} denoting a missing position.

Masks were generated separately for each series. Validation and test masking calls used the deterministic seed offsets $\text{seed}+1$ and $\text{seed}+2$, respectively, and calls with the same input, pattern, rate, and seed reproduce the same mask. Consequently, paired conditions with an unchanged validation or test pattern share the corresponding masked data, and within each condition all candidates receive identical inputs. Mask placement did not depend on series values or labels, and masked original values were not supplied to imputation.

\subsection{Missingness Patterns}
\label{sec:patterns}

\subsubsection{Point Missingness}

Point missingness ($P$) was generated by sampling $k$ distinct time positions uniformly without replacement from the $T$ available positions, so that $|M_P| = k$. Adjacent masked positions were permitted, and each position has marginal inclusion probability $k/T$.

\subsubsection{Circular Block Missingness}

Circular block missingness ($B$) was generated by sampling a starting position $s$ uniformly from $\{0, \ldots, T-1\}$ and masking
\begin{equation}
    M_B = \{(s + j) \bmod T : j = 0, \ldots, k-1\}.
\end{equation}
For example, with $T = 20$ and $k = 5$, a valid zero-based block is $(17, 18, 19, 0, 1)$. Uniform circular placement gives every position marginal inclusion probability $k/T$, so point and circular block masking share the same marginal exposure but differ in the joint arrangement of missing positions.

Circularity applies only to mask construction. Imputation and classification operate on the original linear time axis without assuming periodicity, so a wrapping block appears as missing intervals at both sequence boundaries. The construction serves experimental control rather than modelling a particular real-world failure mechanism.

\subsection{Missingness Rates}
\label{sec:rates}

The six nominal rates were
\begin{equation}
    r \in \{0.05, 0.10, 0.15, 0.20, 0.25, 0.30\}.
\end{equation}
Fixed-count masking removes the same number of observations under both patterns for a given series and rate. Realized rates can differ slightly from nominal rates across datasets because of integer rounding.

When $k = 1$, point and block masking have the same distribution, since each selects a single position uniformly. Their realizations need not coincide, but this case does not compare distinct temporal missingness structures. It occurs only for ItalyPowerDemand ($T = 24$) at the 5\% rate; it was retained in aggregation but is not interpreted as evidence of a structural point-versus-block difference.

\subsection{Imputation}
\label{sec:imputation}

Masked validation and test series were processed by fixed linear interpolation. Missing observations bounded by observed values were interpolated between those endpoints; at either sequence boundary, the nearest observed value was extended to fill the missing interval. Imputation was performed separately within each series using only its observed values, without labels, masked original values, or quantities estimated from other series. The pipeline was therefore
\begin{equation}
    \text{masked series} \rightarrow \text{linear interpolation} \rightarrow \text{classifier}.
\end{equation}
Imputation was not an experimental factor, and the findings are conditional on this procedure.

\subsection{Classification Methods}
\label{sec:classifiers}

The prespecified candidate set consists of three pipelines representing different classification approaches. It is deliberately limited rather than an exhaustive TSC benchmark.

\subsubsection{1NN-DTW}

The distance-based candidate is one-nearest-neighbour classification with dynamic time warping~\cite{bagnall2014nn}. The primary experiment used aeon's \texttt{KNeighbors\allowbreak TimeSeries\allowbreak Classifier} with one neighbour and DTW distance, with no hyperparameter tuning on held-out validation data. The supplementary experiment used a batched prediction wrapper that calls aeon's neighbour-search routine. The wrapper retains the distance calculation and nearest-neighbour tie rule while enabling batched prediction; it is an execution optimization, not a separate candidate method.

\subsubsection{MiniRocket-Ridge}

The convolution-based candidate is aeon's \texttt{MiniRocketClassifier}~\cite{dempster2021minirocket} with 10{,}000 configured kernels and \texttt{RidgeClassifierCV} as its linear classifier. MiniRocket's random state was set to the experimental seed. Ridge regularization was selected internally from
\begin{equation}
    \alpha_j = 10^{-3 + 6j/9}, \qquad j = 0, \ldots, 9,
\end{equation}
using only the complete fitting subset. The grid and fitting rule were fixed, while the selected regularization value could differ across training splits.

\subsubsection{Statistical-Feature Random Forest}

The feature-based candidate extracts ten features from each series: mean, standard deviation, minimum, maximum, median, interquartile range, linear slope, first value, last value, and lag-1 autocorrelation. Training features were computed from the complete fitting series; validation and test features were computed after imputation. The implementation uses the population standard deviation, NumPy~\cite{harris2020array} percentiles for the interquartile range, and the least-squares slope against equally spaced time indices. The lag-1 feature is
\begin{equation}
    a_1 = \frac{\sum_{t=1}^{T-1} (x_t - \bar{x})(x_{t+1} - \bar{x})}{\sum_{t=1}^{T} (x_t - \bar{x})^2},
\end{equation}
with $a_1 = 0$ for a constant series. The feature vectors were classified by a Random Forest~\cite{breiman2001random} with 500 trees and random state equal to the experimental seed.

Candidate definitions, configured hyperparameters, and internal fitting rules were identical across missingness conditions. Settings not explicitly overridden follow the corresponding library defaults and the project implementation.

\subsection{Validation-Based Model Selection}
\label{sec:selection}

For a given dataset and seed, all candidates were fitted on the same complete fitting subset, and selection used balanced accuracy on the masked and imputed validation subset only. Let $\mathcal{C}$ denote the three-candidate set. The maximum validation score is
\begin{equation}
    b^{\mathrm{val}}_{\max} = \max_{j \in \mathcal{C}} \mathrm{BA}^{\mathrm{val}}_j ,
\end{equation}
and candidates within numerical tolerance of it form the tie set
\begin{equation}
    \mathcal{A}_\epsilon = \{ j \in \mathcal{C} : b^{\mathrm{val}}_{\max} - \mathrm{BA}^{\mathrm{val}}_j \le \epsilon \}, \qquad \epsilon = 10^{-12}.
\end{equation}
One candidate was then selected from the name-sorted tie set by a uniform pseudorandom draw. The draw is reproducible: its seed is derived from the dataset, experimental seed, missingness rate, validation pattern, and imputer, and excludes the target pattern and all test outcomes.

A fixed validation condition therefore selects the same candidate under both target patterns, including when scores tie. Tie-set sizes were recorded. Test labels and scores were never used for selection or tuning.

\subsection{Experimental Procedure}
\label{sec:procedure}

For each dataset and seed, the official training set was split into fitting and validation subsets. At each rate, validation and test masks were generated, the masked series were imputed, and all candidates were evaluated. The validation selection rule determined the chosen candidate, whose test score was taken from the recorded candidate results. Test balanced accuracies were retained for all candidates to compute the retrospective oracle defined in \cref{sec:regret}.

The primary runner refitted candidates for each condition and introduced prediction caching during execution. The supplementary runner fitted each candidate once per dataset and seed and reused it across rates and patterns. Both used the same fitting inputs and selection rule. Because the point-side check described in \cref{sec:linear-block} compares results produced by the two runners for every candidate, dataset, seed, and rate, it also shows that these implementation changes did not alter the recorded point-side metrics (see also \cref{sec:implementation}).

For a fixed validation pattern, validation scores are unchanged across the two target patterns; likewise, for a fixed target pattern, each candidate's test score is unchanged across the two validation patterns. These invariances allow corresponding results to be reused and underpin the paired comparisons.

\subsection{Primary Target-Paired Analysis}
\label{sec:target-paired}

The primary comparisons are $PP$ versus $BP$ for point-masked tests and $BB$ versus $PB$ for circular-block-masked tests. Each pair holds fixed the dataset, seed, train--validation split, nominal rate, target test instances and masks, imputation procedure, and candidate methods; only the validation masking-and-selection procedure changes. Individual differences can depend on realized validation masks and tie-breaking outcomes, and are not interpreted as effects of temporal continuity separated from those procedures.

Candidate test scores and the oracle are invariant within each pair. The contrast is
\begin{equation}
    \Delta\mathrm{BA} = \mathrm{BA}_{\mathrm{matched}} - \mathrm{BA}_{\mathrm{mismatched}},
\end{equation}
where each score belongs to the candidate selected under the corresponding validation condition. Positive values indicate worse test performance under mismatched validation; negative values indicate a gain.

\subsection{Secondary Source-Paired Analysis}
\label{sec:source-paired}

Source-paired comparisons fix the validation pattern and change the target pattern: $PP$ versus $PB$, and $BB$ versus $BP$. Validation scores and the validation-only tie-breaking rule are identical within each pair, so the selected candidate is unchanged. These comparisons describe how the selected candidate's test performance and the candidate-set oracle change across target patterns.

Source-paired results are interpreted within each fixed validation-pattern stratum. Their equally weighted overall contrast is not independent corroboration of the primary result, because both overall comparisons assign positive weight to the matched cells ($PP$, $BB$) and negative weight to the mismatched cells ($PB$, $BP$):
\begin{equation}
    \frac{\Delta_P + \Delta_B}{2} = \frac{S_P + S_B}{2},
\end{equation}
where
\begin{align}
    \Delta_P &= \mathrm{BA}_{PP} - \mathrm{BA}_{BP}, & \Delta_B &= \mathrm{BA}_{BB} - \mathrm{BA}_{PB},\\
    S_P &= \mathrm{BA}_{PP} - \mathrm{BA}_{PB}, & S_B &= \mathrm{BA}_{BB} - \mathrm{BA}_{BP}.
\end{align}
The corresponding identity holds for the overall selection-error contrast. These identities concern the complete, equally weighted design, not subsets with different retained observations or weights.

\subsection{Evaluation Metrics}
\label{sec:metrics}

\subsubsection{Balanced Accuracy}

Balanced accuracy, which averages recall over classes and is therefore not inflated by class imbalance~\cite{brodersen2010balanced}, was computed using scikit-learn~\cite{JMLR:v12:pedregosa11a}, specifically its \texttt{balanced\_accuracy\_score} implementation. For the class set $\mathcal{C}_y$ represented in the evaluated ground-truth labels,
\begin{equation}
    \mathrm{BA} = \frac{1}{|\mathcal{C}_y|} \sum_{c \in \mathcal{C}_y} \frac{\mathrm{TP}_c}{\mathrm{TP}_c + \mathrm{FN}_c}.
\end{equation}
Balanced accuracy was used for selection, oracle identification, and the primary performance analysis. Macro-F1 was recorded as a secondary descriptive metric only.

\subsubsection{Mismatch Effect}
\label{sec:mismatch-effect}

The target-specific primary effects are
\begin{equation}
    \Delta\mathrm{BA}_P = \mathrm{BA}_{PP} - \mathrm{BA}_{BP}, \qquad
    \Delta\mathrm{BA}_B = \mathrm{BA}_{BB} - \mathrm{BA}_{PB}.
\end{equation}
Each compares validation-based choices on the same masked target test set rather than comparing different test patterns directly.

\subsubsection{Selection Regret}
\label{sec:regret}

The candidate-set test oracle is the highest observed test balanced accuracy among the three prespecified candidates,
\begin{equation}
    \mathrm{BA}^{\mathrm{oracle}} = \max_{j \in \mathcal{C}} \mathrm{BA}^{\mathrm{test}}_j .
\end{equation}
It is a retrospective evaluation reference only. Candidates within $10^{-12}$ of the maximum were retained as oracle members, while the oracle score remained the maximum recorded score. Regret is
\begin{equation}
    R = \mathrm{BA}^{\mathrm{oracle}} - \mathrm{BA}^{\mathrm{selected}},
\end{equation}
and within a target pair
\begin{equation}
    \Delta\mathrm{BA} = R_{\mathrm{mismatched}} - R_{\mathrm{matched}}.
\end{equation}
The regret difference is therefore an equivalent representation of the primary effect, while absolute regret describes each choice's distance from the observed candidate-set maximum.

\subsubsection{Selection Error}

Selection error is the indicator
\begin{equation}
    \mathbb{I}\left(\mathrm{BA}^{\mathrm{selected}} < \mathrm{BA}^{\mathrm{oracle}} - 10^{-12}\right).
\end{equation}
It denotes observed suboptimality within the candidate set beyond numerical tolerance, not a statistically established difference in population performance.

\subsubsection{Selection Change}

Selection change records whether the selected candidate differs between the matched and mismatched validation conditions. For a fixed dataset, seed, and rate, the indicator is identical in both target strata because selection depends only on validation. Its frequencies are reported descriptively and are not interpreted as performance loss.

\subsection{Dataset-Level Aggregation and Statistical Analysis}
\label{sec:statistics}

Datasets were treated as the independent analysis units; seeds, rates, and patterns are repeated measurements within datasets. Let $\Delta_{d,s,r,t}$ denote the target-paired effect for dataset $d$, seed $s$, rate $r$, and target pattern $t$. The dataset-level effect is
\begin{equation}
    \bar{\Delta}_d = \frac{1}{|\mathcal{S}|\,|\mathcal{R}|\,|\mathcal{T}|} \sum_{s \in \mathcal{S}} \sum_{r \in \mathcal{R}} \sum_{t \in \mathcal{T}} \Delta_{d,s,r,t},
\end{equation}
where $\mathcal{S} = \{1, \ldots, 5\}$ is the set of seeds, $\mathcal{R} = \{0.05, 0.10, \ldots, 0.30\}$ the set of nominal missingness rates, and $\mathcal{T} = \{P, B\}$ the set of target patterns, so that $|\mathcal{S}|\,|\mathcal{R}|\,|\mathcal{T}| = 5 \times 6 \times 2 = 60$. The overall estimate is
\begin{equation}
    \hat{\Delta} = \frac{1}{64} \sum_{d=1}^{64} \bar{\Delta}_d .
\end{equation}
This weights datasets and the evaluated seeds, rates, and target patterns equally, rather than by test-set size or deployment frequency. Target-specific analyses average over seeds and rates; rate-specific analyses average over seeds and target patterns.

Signed contrasts were summarized by the mean, median, direction counts, a 95\% percentile bootstrap interval for the mean~\cite{efron1993bootstrap}, and a two-sided Wilcoxon signed-rank test~\cite{JMLR:v7:demsar06a}. Values within $10^{-9}$ of zero were counted as no difference in the direction counts; this tolerance is separate from the $10^{-12}$ selection tolerance.

Bootstrap intervals used 10{,}000 resamples of datasets and analysis seed 20260928. They describe variation under resampling of the included tasks, treating datasets as exchangeable resampling units, rather than probability-sampling uncertainty for all TSC tasks; test instances and mask realizations were not separately resampled.

Wilcoxon tests were computed with SciPy~\cite{virtanen2020scipy} on the dataset-level values with \texttt{zero\_method=\allowbreak"wilcox"}, which omits exactly zero differences rather than splitting their ranks between the positive and negative sums as described by Dem\v{s}ar~\cite{JMLR:v7:demsar06a}. The direction-count tolerance does not recode the test inputs; when all dataset-level values lie within that tolerance of zero, the implementation reports $p = 1$. The signed-rank test is not specifically a test of the arithmetic mean, and a location-shift interpretation requires the usual assumptions about the difference distribution.

For each signed metric in the rate-specific analysis, the six rate-specific $p$-values were Holm-adjusted~\cite{holm1979simple}. This is not a joint correction across the overall, stratum-specific, trend, and supplementary analyses. Selection-change frequencies are descriptive; generic signed-rank output for this nonnegative quantity is not used as evidence of performance harm.

\subsection{Missingness-Rate Trend and Sensitivity Analyses}
\label{sec:sensitivity}

For each dataset, effects were averaged over seeds and target patterns at each rate, and a least-squares line was fitted across the six rates:
\begin{equation}
    \Delta(r) = \beta_0 + \beta_1 (100 r).
\end{equation}
The slope $\beta_1$ is in balanced-accuracy units per percentage point of nominal missingness; multiplying it by 100 expresses it in balanced-accuracy percentage points per percentage point. Dataset-level slopes were summarized with the same bootstrap and signed-rank framework.

A tie-exclusion sensitivity analysis removed target pairs containing a validation tie on either side. Dataset-level means were recomputed from the remaining pairs, and datasets without remaining pairs were omitted. This conditions the summary on tie-free observations rather than reproducing the original aggregation.

Finite-count stratified splits can omit a class from validation. Such splits were retained; their validation scores provide no recall estimate for the absent class. The $k = 1$ condition for ItalyPowerDemand at 5\% was likewise retained without interpreting it as a structural point-versus-block comparison.

The five seeds jointly vary the split, the applicable model random states, and mask generation; they do not provide separate estimates of uncertainty from these sources.

\subsection{Supplementary Linear-Block Robustness Analysis}
\label{sec:linear-block}

The supplementary experiment replaced circular masks with non-wrapping linear blocks,
\begin{equation}
    M_L = \{s, s+1, \ldots, s+k-1\}, \qquad s \sim \mathrm{Uniform}\{0, \ldots, T-k\}.
\end{equation}
The block remains on the original linear time axis. For $1 < k < T$, boundary positions generally have lower inclusion probabilities than interior positions; this exposure difference is part of the construction being assessed.

The supplementary analysis used the same 64 datasets, five seeds, six rates, splitting procedure, candidate definitions and hyperparameters, interpolation, selection rule, metrics, and dataset-level framework. It newly evaluated three conditions: $PL$ (point-masked validation, linear-block-masked test), $LP$ (linear-block-masked validation, point-masked test), and $LL$ (linear-block masking in both splits). These produced
\begin{equation}
    64 \times 5 \times 6 \times 3 \times 3 = 17{,}280
\end{equation}
new classifier-condition records.

The primary $PP$ results were reused, because this condition involves no block masking and is unaffected by the alternative construction. Within a dataset, seed, and rate, $LP$ and $LL$ shared the linear-block validation mask, and $PL$ and $LL$ shared the linear-block test mask; point-side rules and seed offsets were unchanged. Before $PP$ was reused, the point-side validation and test balanced accuracies of the supplementary run were checked against the corresponding primary values for every candidate, dataset, seed, and rate. This verification concerned recorded metrics, not direct comparison of mask arrays, prediction arrays, or estimator objects.

The supplementary target pairs were $PP$ versus $LP$ for point-masked tests and $LL$ versus $PL$ for linear-block-masked tests, analysed with the same aggregation framework with $L$ in place of the block target stratum. Primary and supplementary results are reported separately and never pooled. The supplementary experiment assesses whether the primary finding also holds under another block construction; it does not test for an inferential difference between the constructions.

The supplementary run used 24 parallel jobs instead of the 18 used in the primary run. Parallelism is an implementation setting, not an experimental factor.

\subsection{Experimental Integrity and Reproducibility}
\label{sec:integrity}

The study maintained complete fitting data, validation-only selection, identical candidate inputs within each condition, separate validation and test mask generation, and within-series imputation from observed values only.

Before analysis, selection, oracle membership, selection error, and regret were reconstructed from the classifier-level results. Automated checks covered duplicate keys, candidate membership, missing values, result reconstruction, and target-paired oracle and regret consistency. Result-verification notes additionally document configuration coverage, masking counts, metric ranges, and the expected source--target invariances.

Configurations, raw metric files, analysis scripts, and verification notes are retained with the project, and configuration and assembly manifests identify the analysis inputs. The recorded metrics permit reconstruction of selection outcomes but do not substitute for archived per-instance predictions or fitted model objects. Timing fields that repeat reused fits or predictions are not additive wall-clock costs.

Inference is conditional on the selected benchmark tasks, candidate set, interpolation procedure, and implemented randomization. Related tasks may be dependent, and dataset-level resampling does not explicitly model task-family dependence. The conclusions therefore concern the studied controlled setting rather than unrestricted TSC or deployment scenarios.

\section{Experimental Setup}
\label{sec:experimental-setup}

\subsection{Dataset Configuration}
\label{sec:dataset-configuration}

Both experiments used the fixed, ordered list of 64 UCR2015 datasets defined in \cref{sec:datasets}, stored in the primary configuration; no dataset was added or removed after the list was fixed.

\subsection{Formal Experimental Runs}
\label{sec:formal-runs}

The primary experiment used the configuration \nolinkurl{configs/main_protocol_balanced.yaml}. With 64 datasets, five seeds, six missingness rates, four validation--test pattern conditions, and three candidate classifiers, it produced
\begin{equation}
    64 \times 5 \times 6 \times 4 \times 3 = 23{,}040
\end{equation}
classifier-condition records and
\begin{equation}
    64 \times 5 \times 6 \times 4 = 7{,}680
\end{equation}
selection-condition records.

The supplementary experiment used \nolinkurl{configs/supplementary_linear_block.yaml} and evaluated the three linear-block conditions defined in \cref{sec:linear-block}, producing 17{,}280 classifier-condition records and 5{,}760 selection-condition records. Adding the reused primary $PP$ records gave an assembled supplementary input of 23{,}040 classifier-condition and 7{,}680 selection-condition records. Both formal result sets contain complete results for all 64 configured datasets.

The official UCR test sequences and labels were retained as the target evaluation set. Missingness was applied to the test input sequences according to the target condition, and the test labels were used only to evaluate the candidates and compute the retrospective oracle, never for selection.

\subsection{Implementation}
\label{sec:implementation}

The experiments were implemented in Python~3.12 and run with aeon~1.6.0 on a single Windows~11 workstation with an AMD Ryzen AI~9 HX~370 processor (12 cores, 24 logical processors) and 32\,GB of memory. The primary run used 18 parallel jobs and the supplementary run 24; the supplementary run took about 3.7~hours of wall-clock time. The supported version ranges of aeon, scikit-learn, NumPy, SciPy, and pandas~\cite{mckinney2010data} are recorded in the project's requirements file. Interpolation, candidate definitions, hyperparameters, random-state handling, and validation-based selection followed \cref{sec:imputation,sec:classifiers,sec:selection}.

The primary results were generated by the original runner, which fitted the candidates separately for each condition; prediction caching was introduced during this run, after the first nine datasets. The supplementary run used a later, optimized runner that fits each candidate once per dataset and seed, reuses the fitted estimator across rates and pattern conditions, and predicts 1NN-DTW queries in batches. In both runners, the fitting data, candidate specifications, and validation-based selection rule were unchanged; the differences concern computational execution rather than the experimental design or estimands.

Three checks support the equivalence of these execution paths. First, a regression test compares the batched 1NN-DTW predictions with those of aeon's reference \texttt{KNeighbors\allowbreak TimeSeries\allowbreak Classifier}, including cases with tied distances. Second, the first nine primary datasets, produced before prediction caching was added, satisfy the same validation- and test-score invariances that caching relies on. Third, the point-side comparison described in \cref{sec:analysis-execution} matches outputs of the two runners for every candidate, dataset, seed, and rate.

\subsection{Analysis Execution and Verification}
\label{sec:analysis-execution}

The frozen classifier-level and selection-level result files were analysed with \texttt{paired\_analysis.py}, using the aggregation, inferential, trend, and sensitivity procedures defined in \cref{sec:statistics,sec:sensitivity}. The formal analyses used 10{,}000 dataset-level bootstrap resamples with analysis seed 20260928. The figures were produced by \texttt{plot\_final\_results.py} from the same result files, with intervals obtained by the same dataset-level bootstrap.

Before any summary statistic was produced, the analysis program checked the structure and internal consistency of the archived results. These checks covered duplicate and missing entries, expected candidate membership, dataset completeness, and reconstruction of the stored selection quantities from the classifier-level records. Construction of the target pairs additionally checked the shared test oracle and the numerical agreement between the balanced-accuracy and regret representations of the paired effect.

A separate archive-level verification, distinct from the checks executed automatically by the analysis program, covered record counts, configuration coverage, metric ranges, masking counts, and the validation- and test-score invariances implied by the experimental design.

For the supplementary analysis, the primary $PP$ records were reused only after the point-side validation and test balanced accuracies of the supplementary run had been compared with the corresponding primary values. They agreed within the assembly tolerance of $10^{-12}$ for all 64 datasets before the combined supplementary analysis files were created. The primary and supplementary analyses were conducted separately and were not pooled.

The configuration files, frozen result files, analysis outputs, and verification records constitute the numerical record of the reported study. Execution history, file checksums, and detailed verification notes are documented alongside the result files.

\section{Results}
\label{sec:results}

All results in this section are dataset-level summaries over the 64 datasets (\cref{sec:statistics}). Differences in balanced accuracy are reported in percentage points (pp). Unless stated otherwise, intervals are 95\% percentile bootstrap intervals, direction counts give the number of datasets with a worse, better, or unchanged outcome under mismatched validation, and $p$-values are from two-sided Wilcoxon signed-rank tests. Sections~\ref{sec:primary-effect}--\ref{sec:sensitivity-results} concern the primary circular-block experiment; \cref{sec:linear-block-results} reports the supplementary linear-block analysis separately.

\subsection{Primary Target-Paired Effect}
\label{sec:primary-effect}

Mismatched validation reduced the test balanced accuracy of the selected candidate (\cref{tab:primary}). Averaged over seeds, rates, and both target patterns, the loss was 1.14~pp (95\% CI 0.79 to 1.51; median 0.55~pp). It was positive for 49 datasets, negative for 7, and zero for 8 ($p = 1.8 \times 10^{-9}$). Because the oracle is shared within each target pair, the same value is the mean increase in selection regret. Mismatch also increased the selection-error rate by 6.4~pp (3.6 to 9.4; 28/8/28 datasets; $p = 1.7 \times 10^{-5}$), and the selected candidate changed in 35.5\% of target pairs (30.1\% to 40.9\%).

The effect differed between the two target patterns. For point-masked tests, selecting under block validation ($BP$) instead of point validation ($PP$) lowered the selected candidate's test balanced accuracy by 1.84~pp (1.23 to 2.50; 43/13/8 datasets; $p = 7.4 \times 10^{-7}$) and raised the selection-error rate by 11.5~pp (7.1 to 16.2; $p = 2.5 \times 10^{-5}$). For circular-block-masked tests, selecting under point validation ($PB$) instead of block validation ($BB$) lowered it by 0.44~pp, with an interval that included zero ($-0.11$ to 1.03; 34/22/8 datasets; $p = 0.20$); the change in selection error was likewise small and not significant (1.4~pp, $-2.3$ to 5.1; $p = 0.46$). The overall effect is therefore driven mainly by point-masked deployment.

\begin{table}[t]
\centering
\caption{Target-paired effect of mismatched validation in the primary experiment. $\Delta$BA is the loss in the selected candidate's test balanced accuracy (equivalently, the increase in regret); $\Delta$SE is the increase in the selection-error rate. Means over 64 datasets with 95\% bootstrap intervals; W/B/E counts datasets on which mismatch was worse, better, or equal; $p$ from two-sided Wilcoxon signed-rank tests. The tie-excluded row omits target pairs with a validation tie on either side (62 datasets remain).}
\label{tab:primary}
\small
\begin{tabular}{@{}l r c c r r c r@{}}
\toprule
& \multicolumn{4}{c}{$\Delta$BA (pp)} & \multicolumn{3}{c}{$\Delta$SE (pp)} \\
\cmidrule(lr){2-5} \cmidrule(l){6-8}
Comparison & Mean & 95\% CI & W/B/E & $p$ & Mean & 95\% CI & $p$ \\
\midrule
Both targets          & 1.14 & [0.79, 1.51]  & 49/7/8  & $1.8\times10^{-9}$ & 6.4  & [3.6, 9.4]   & $1.7\times10^{-5}$ \\
Point test ($BP$ vs $PP$) & 1.84 & [1.23, 2.50]  & 43/13/8 & $7.4\times10^{-7}$ & 11.5 & [7.1, 16.2]  & $2.5\times10^{-5}$ \\
Block test ($PB$ vs $BB$) & 0.44 & [$-0.11$, 1.03] & 34/22/8 & 0.20             & 1.4  & [$-2.3$, 5.1] & 0.46 \\
\midrule
Both targets, ties excluded & 1.23 & [0.75, 1.78] & 44/6/12 & $5.3\times10^{-8}$ & 8.5 & [4.5, 13.3] & $6.5\times10^{-5}$ \\
\bottomrule
\end{tabular}
\end{table}

\subsection{Dependence on Missingness Rate}
\label{sec:rate-results}

\begin{figure}[t]
\centering
\includegraphics[width=\textwidth]{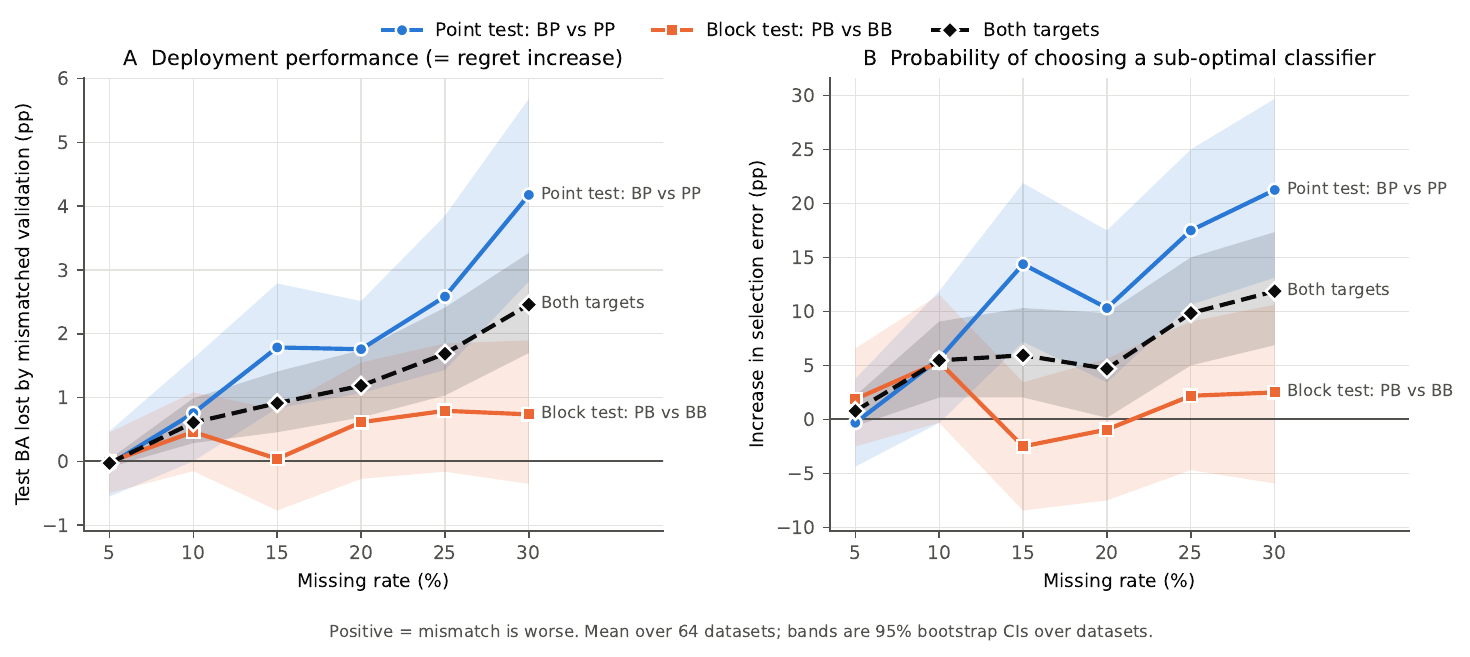}
\caption{Target-paired cost of mismatched validation by missingness rate in the primary experiment. (A)~Loss in the selected candidate's test balanced accuracy, equal to the increase in regret. (B)~Increase in the selection-error rate. Lines show means over 64 datasets for point-masked tests ($BP$ vs $PP$), circular-block-masked tests ($PB$ vs $BB$), and both targets; bands are 95\% bootstrap intervals over datasets. Positive values indicate that mismatch is worse.}
\label{fig:rate}
\end{figure}

The cost of mismatch grew with the missingness rate (\cref{tab:rates,fig:rate}). At 5\% it was indistinguishable from zero ($-0.03$~pp, $-0.09$ to 0.03; Holm-adjusted $p = 0.79$), rising to 0.61~pp at 10\% and 2.46~pp at 30\% (1.70 to 3.27). From 10\% upward, every rate-specific mean effect was positive and statistically significant after Holm adjustment across the six rates. The dataset-level least-squares slope was 0.091~pp of balanced accuracy per percentage point of missingness (0.061 to 0.124; 48/8/8 datasets with a positive, negative, or zero slope; $p = 6.4 \times 10^{-8}$). The increase was again concentrated in point-masked tests, where the loss reached 4.2~pp at 30\%, whereas the block-test curve remained below 1~pp at every rate (\cref{fig:rate}A).

The selected candidate changed more often as the rate increased, from 21.3\% of target pairs at 5\% to 48.1\% at 30\%. At 5\%, mismatch therefore altered the selection in about one pair in five without a detectable loss in test performance: at low rates the two validation patterns usually led to candidates that performed similarly on the target test set. The selection-error increase was less regular across rates, being significant after Holm adjustment at 10\%, 15\%, 25\%, and 30\% but not at 5\% or 20\% (\cref{tab:rates}).

\begin{table}[t]
\centering
\caption{Target-paired effect by nominal missingness rate in the primary experiment, averaged over seeds and both target patterns. $p_{\mathrm{Holm}}$ values are Holm-adjusted across the six rates within each metric. \emph{Changed} is the percentage of target pairs in which mismatched validation selected a different candidate.}
\label{tab:rates}
\small
\begin{tabular}{@{}r r c c c r c r@{}}
\toprule
& \multicolumn{4}{c}{$\Delta$BA (pp)} & \multicolumn{2}{c}{$\Delta$SE (pp)} & \\
\cmidrule(lr){2-5} \cmidrule(lr){6-7}
Rate & Mean & 95\% CI & W/B/E & $p_{\mathrm{Holm}}$ & Mean & $p_{\mathrm{Holm}}$ & Changed (\%) \\
\midrule
5\%  & $-0.03$ & [$-0.09$, 0.03] & 18/17/29 & 0.79                & 0.8  & 0.27                & 21.3 \\
10\% & 0.61    & [0.29, 0.99]    & 30/12/22 & $3.0\times10^{-3}$ & 5.5  & 0.018               & 27.8 \\
15\% & 0.91    & [0.46, 1.41]    & 34/12/18 & $2.0\times10^{-4}$ & 5.9  & 0.037               & 35.3 \\
20\% & 1.19    & [0.69, 1.74]    & 40/10/14 & $2.8\times10^{-5}$ & 4.7  & 0.26                & 36.9 \\
25\% & 1.69    & [1.03, 2.41]    & 40/11/13 & $9.5\times10^{-6}$ & 9.8  & $1.8\times10^{-3}$ & 43.8 \\
30\% & 2.46    & [1.70, 3.27]    & 47/9/8   & $7.1\times10^{-8}$ & 11.9 & $2.6\times10^{-4}$ & 48.1 \\
\bottomrule
\end{tabular}
\end{table}

\subsection{Cross-Pattern Generalization}
\label{sec:cross-pattern}

\begin{figure}[t]
\centering
\includegraphics[width=\textwidth]{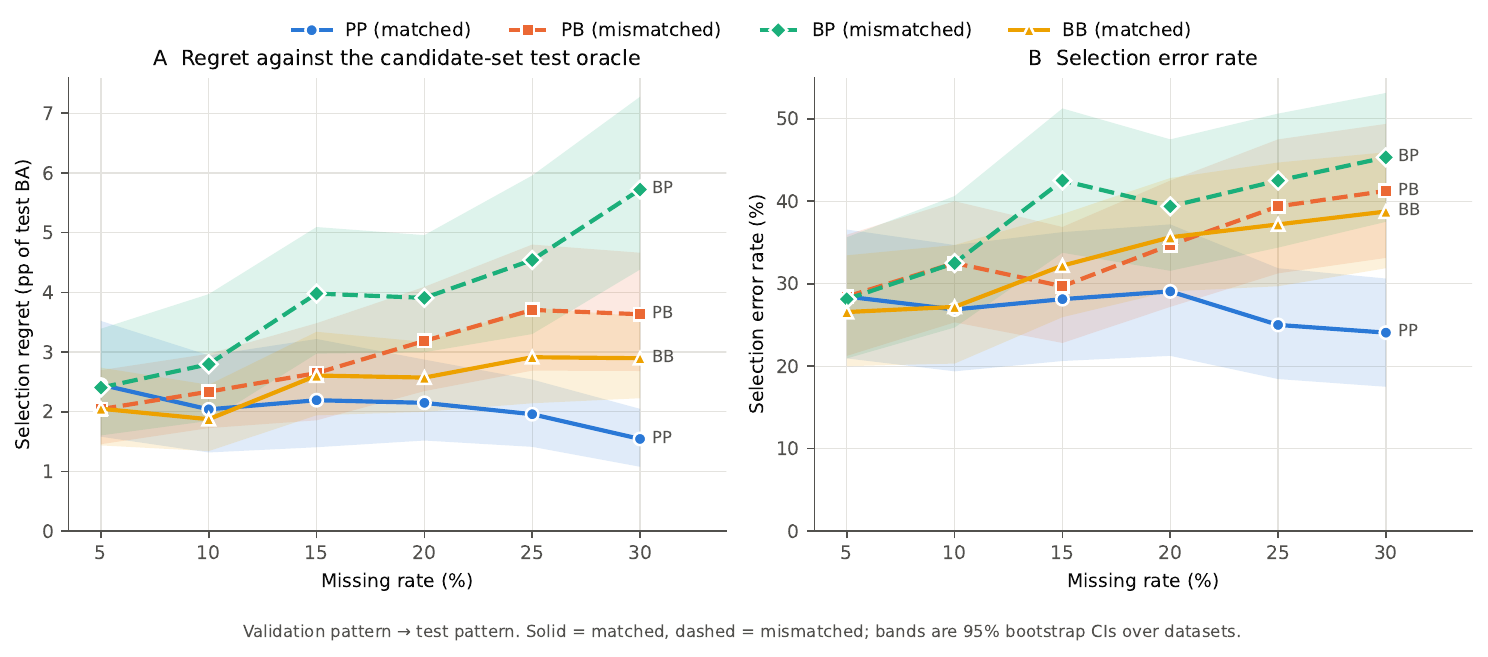}
\caption{Outcomes of the four primary conditions by missingness rate. (A)~Regret against the candidate-set test oracle. (B)~Selection-error rate. Solid lines are matched and dashed lines mismatched conditions; bands are 95\% bootstrap intervals over 64 datasets.}
\label{fig:conditions}
\end{figure}

\Cref{tab:conditions,fig:conditions} describe the four conditions separately. Circular block masking of the test set was considerably harder than point masking: the candidate-set oracle averaged 81.5\% balanced accuracy on point-masked tests and 70.4\% on block-masked tests, a difference of 11.1~pp, with the block-test oracle lower on 62 of the 64 datasets. Matched validation did not remove the effect of selection error. Even in $PP$, the selected candidate was suboptimal in 26.9\% of conditions and fell 2.06~pp short of the oracle on average, reflecting the limited size of the validation subsets.

Within each test pattern, the matched condition had the lower regret and selection-error rate. Regret rose from 2.06~pp ($PP$) to 3.89~pp ($BP$) on point-masked tests and from 2.49~pp ($BB$) to 2.93~pp ($PB$) on block-masked tests. Across rates (\cref{fig:conditions}), the regret of $PP$ stayed near 2~pp or decreased slightly, whereas that of $BP$ rose to 5.7~pp at 30\%; the two block-test conditions remained close to each other.

The secondary source-paired comparisons fix the validation pattern and change the target pattern (\cref{sec:source-paired}). With point validation, moving the test from point to block masking ($PP$ to $PB$) increased regret by 0.87~pp (0.25 to 1.51; $p = 3.7 \times 10^{-3}$); with block validation, moving the test from block to point masking ($BB$ to $BP$) increased it by 1.41~pp (0.74 to 2.17; $p = 7.2 \times 10^{-5}$). In both strata, regret was larger when the test pattern differed from the validation pattern, so validation choices did not transfer without loss to the other test pattern. As shown in \cref{sec:source-paired}, the equally weighted average of these two contrasts is algebraically identical to the overall target-paired effect (1.14~pp) and does not constitute independent evidence.

\begin{table}[t]
\centering
\caption{Dataset-level outcomes of the four primary conditions (validation pattern $\to$ test pattern), averaged over seeds and rates; 95\% bootstrap intervals in brackets. The oracle depends only on the test pattern and the tie rate only on the validation pattern.}
\label{tab:conditions}
\small
\begin{tabular}{@{}l c c c c c@{}}
\toprule
& Regret & Selection error & Selected BA & Oracle BA & Validation ties \\
Condition & (pp) & (\%) & (\%) & (\%) & (\%) \\
\midrule
$PP$ (matched)    & 2.06 [1.49, 2.71] & 26.9 [20.5, 33.5] & 79.4 & 81.5 & 20.1 \\
$BP$ (mismatched) & 3.89 [3.03, 4.89] & 38.4 [31.7, 45.2] & 77.6 & 81.5 & 13.4 \\
$BB$ (matched)    & 2.49 [2.01, 2.99] & 32.9 [27.7, 38.3] & 67.9 & 70.4 & 13.4 \\
$PB$ (mismatched) & 2.93 [2.27, 3.60] & 34.3 [28.2, 40.4] & 67.4 & 70.4 & 20.1 \\
\bottomrule
\end{tabular}
\end{table}

\subsection{Classifier Selection}
\label{sec:classifier-results}

\begin{figure}[t]
\centering
\includegraphics[width=\textwidth]{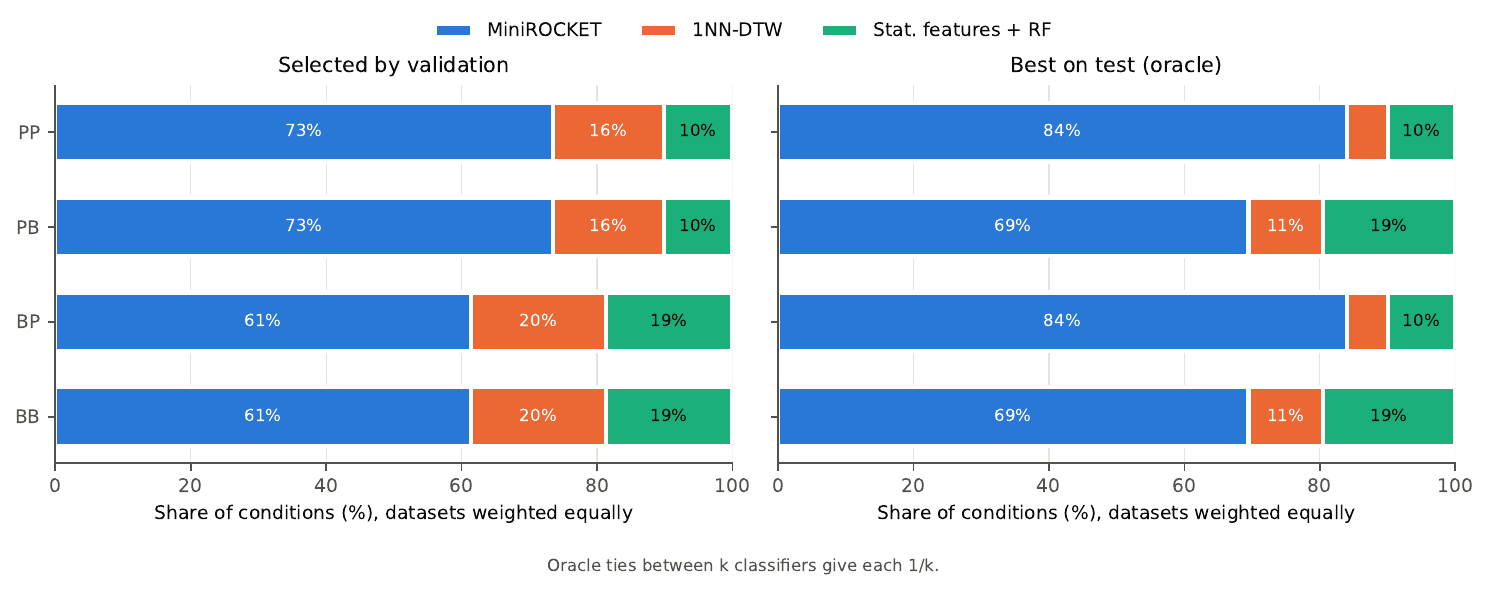}
\caption{Candidate shares in the primary experiment, weighting datasets equally. Left: candidate selected by validation balanced accuracy. Right: best candidate on the test set (oracle); ties between $k$ candidates give each a share of $1/k$. Selection depends only on the validation pattern and the oracle only on the test pattern.}
\label{fig:classifiers}
\end{figure}

The validation pattern changed which candidates were selected (\cref{fig:classifiers}). Under point validation, MiniRocket-Ridge was selected in 73.5\% of conditions, 1NN-DTW in 16.4\%, and the statistical-feature Random Forest in 10.2\%. Under block validation, the share of MiniRocket-Ridge fell to 61.4\%, while 1NN-DTW and the Random Forest rose to 19.9\% and 18.7\%. MiniRocket-Ridge was selected at least once on every dataset under both validation patterns; 1NN-DTW and the Random Forest were selected on 51 datasets each under block validation, compared with 37 and 32 under point validation.

The test oracle showed a different structure. On point-masked tests, MiniRocket-Ridge accounted for 84.0\% of the fractional oracle share; on block-masked tests its share fell to 69.5\%, while the Random Forest and 1NN-DTW accounted for 19.5\% and 11.0\%, respectively. The shift towards 1NN-DTW and the Random Forest under block validation was therefore costly on point-masked tests. When selected in $BP$, these two candidates were on average 10.5~pp and 14.5~pp below the oracle, compared with 0.5~pp for MiniRocket-Ridge. Conversely, point validation concentrated selection on MiniRocket-Ridge, which remained competitive on block-masked tests (1.97~pp below the oracle when selected in $PB$); this is consistent with the smaller block-test effect.

A changed selection did not always reduce performance. Pooled over the 1{,}364 target pairs in which the selected candidate changed, the mismatched selection was worse in 58.3\%, better in 37.3\%, and equal in test balanced accuracy in the remaining 4.4\%, with a mean loss of 3.2~pp. Classifier identity and test performance therefore carry different information, as anticipated in \cref{sec:related}.

\subsection{Dataset-Level Heterogeneity}
\label{sec:heterogeneity}

\begin{figure}[tbp]
\centering
\includegraphics[width=0.72\textwidth]{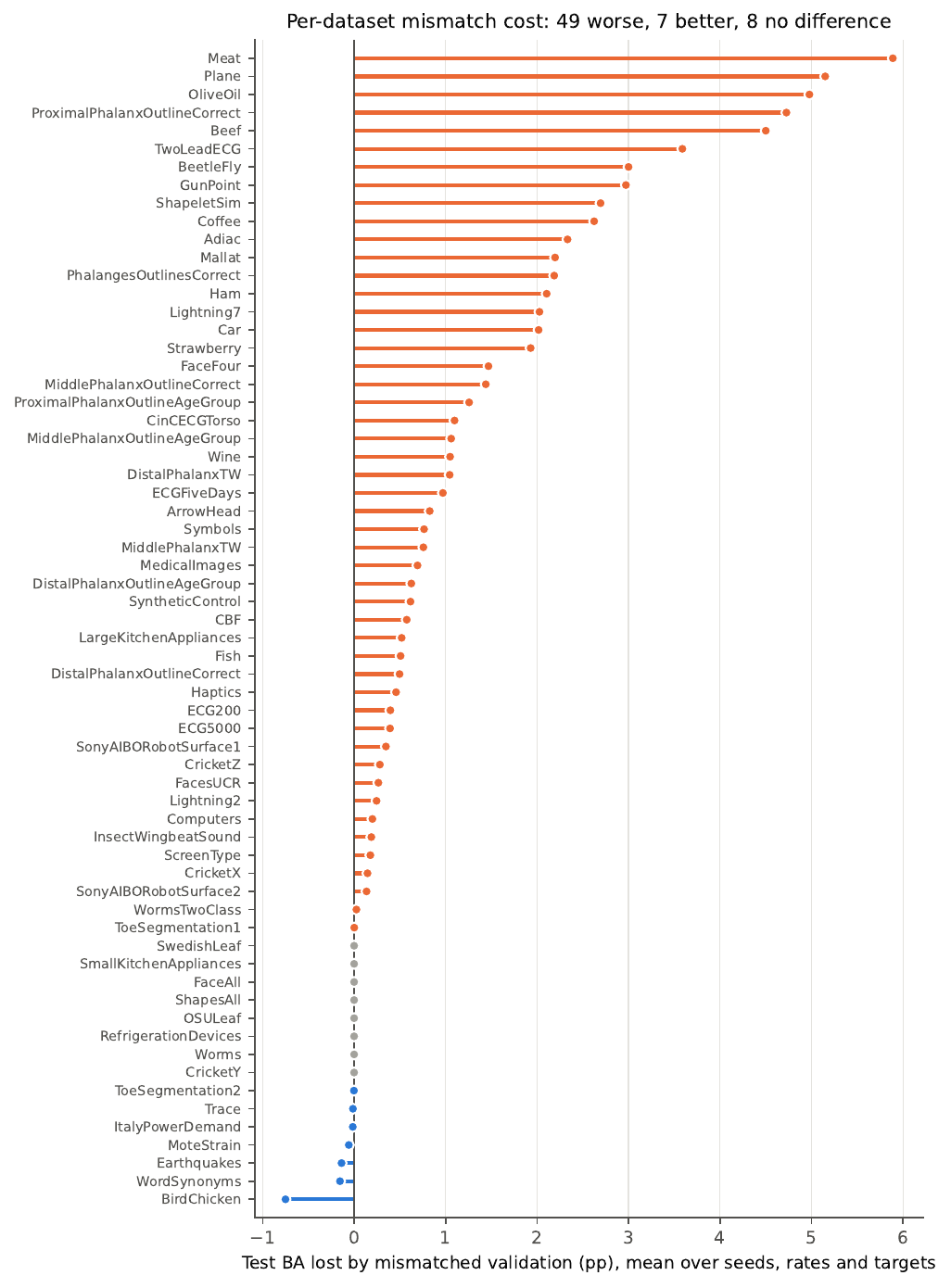}
\caption{Dataset-level target-paired cost of mismatched validation in the primary experiment: loss in the selected candidate's test balanced accuracy, averaged over seeds, rates, and both target patterns. Orange: mismatch worse; blue: mismatch better; grey: no difference within $10^{-9}$.}
\label{fig:datasets}
\end{figure}

The dataset-level effects were predominantly small and positive (\cref{fig:datasets}). Twenty-four datasets lost more than 1~pp under mismatched validation, and the largest losses were observed for Meat (5.89~pp), Plane (5.15~pp), OliveOil (4.98~pp), ProximalPhalanxOutlineCorrect (4.73~pp), and Beef (4.50~pp). The seven datasets with a gain under mismatch all had gains below 1~pp, the largest being BirdChicken ($-0.75$~pp). Three of these five datasets (Beef, Meat, and OliveOil) have at most 60 training instances, so their validation subsets are small. BirdChicken, with only 20 training instances, is also small, which suggests that small validation subsets make dataset-level effects more variable in both directions; this observation is descriptive and was not tested.

\subsection{Sensitivity to Validation Ties}
\label{sec:sensitivity-results}

Validation ties occurred in 20.1\% of conditions under point validation and in 13.4\% under block validation. Excluding all target pairs with a tie on either side left 62 datasets and did not weaken the result: the overall loss was 1.23~pp (0.75 to 1.78; 44/6/12 datasets; $p = 5.3 \times 10^{-8}$), and the selection-error increase was 8.5~pp (\cref{tab:primary}). The primary effect is therefore not an artefact of the random tie-breaking rule.

\subsection{Supplementary Linear-Block Analysis}
\label{sec:linear-block-results}

\begin{figure}[t]
\centering
\includegraphics[width=\textwidth]{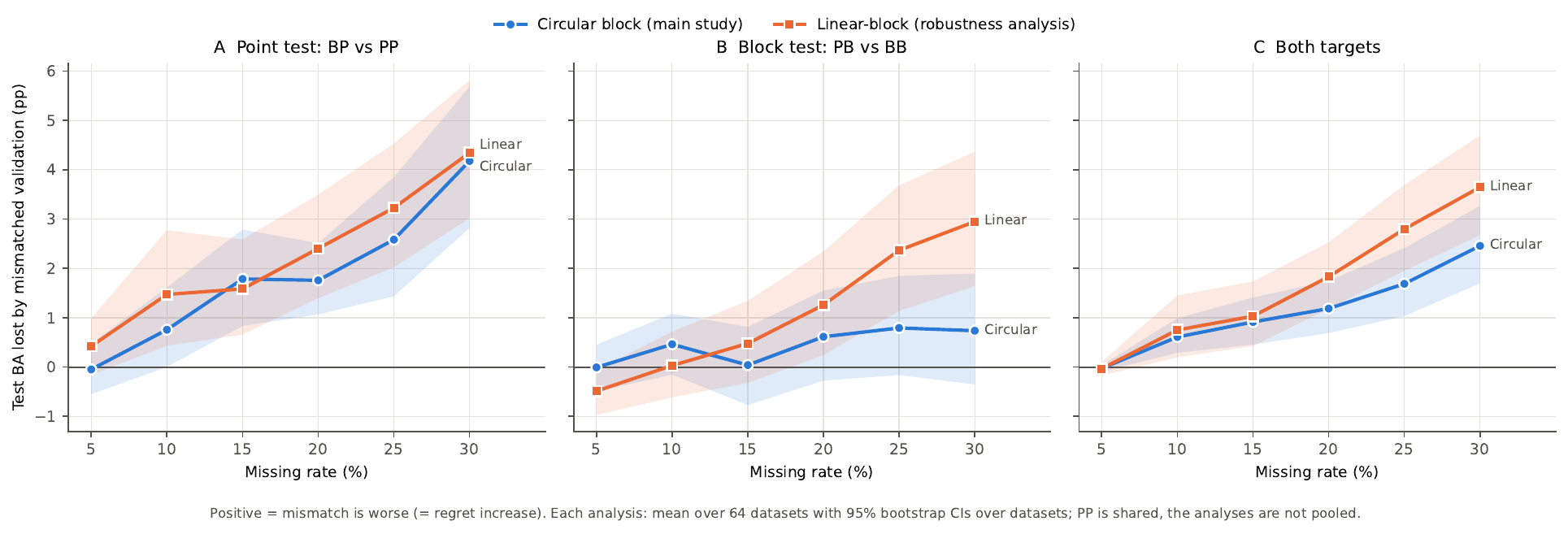}
\caption{Target-paired cost of mismatched validation under circular blocks (primary experiment) and linear blocks (supplementary experiment) by missingness rate: (A)~point-masked tests, (B)~block-masked tests, (C)~both targets. Each analysis is summarized separately over 64 datasets with 95\% bootstrap intervals; the $PP$ condition is shared, and the analyses are not pooled.}
\label{fig:linear}
\end{figure}

Replacing circular blocks with non-wrapping linear blocks (\cref{sec:linear-block}) reproduced the main target-paired pattern and yielded a larger overall balanced-accuracy effect (\cref{tab:linear,fig:linear}). The overall loss was 1.67~pp (1.23 to 2.14; 54/4/6 datasets; $p = 1.2 \times 10^{-10}$), and the selection-error increase was 11.2~pp (7.3 to 15.4; $p = 1.2 \times 10^{-7}$). For point-masked tests ($LP$ vs $PP$), the loss was 2.24~pp, close to the 1.84~pp of the primary experiment. For block-masked tests ($PL$ vs $LL$), the loss was 1.10~pp (0.45 to 1.81; $p = 0.013$), and unlike in the primary experiment its interval excluded zero. The effect again increased with the rate, from $-0.03$~pp at 5\% to 3.65~pp at 30\%, with a slope of 0.145~pp per percentage point (0.100 to 0.192); rate-specific effects were significant after Holm adjustment from 10\% upward. Excluding validation ties left 62 datasets and an overall loss of 1.93~pp (1.26 to 2.69). Mean regret was 2.06~pp for $PP$, 2.43~pp for $LL$, 3.53~pp for $PL$, and 4.30~pp for $LP$.

\Cref{fig:linear} shows that the two constructions differ mainly for block-masked tests. As specified in \cref{sec:linear-block}, the two analyses are summarized separately and are not pooled or tested against each other.

\begin{table}[t]
\centering
\caption{Target-paired effect of mismatched validation under circular blocks (primary experiment) and non-wrapping linear blocks (supplementary experiment). Each analysis uses 64 datasets and is summarized separately; the $PP$ condition is shared. Values are means with 95\% bootstrap intervals; the trend is the mean dataset-level slope in pp of balanced accuracy per percentage point of missingness.}
\label{tab:linear}
\small
\begin{tabular}{@{}l c c@{}}
\toprule
Outcome & Circular block & Linear block \\
\midrule
$\Delta$BA, both targets (pp)            & 1.14 [0.79, 1.51]   & 1.67 [1.23, 2.14] \\
\quad W/B/E datasets; $p$                 & 49/7/8; $1.8\times10^{-9}$ & 54/4/6; $1.2\times10^{-10}$ \\
$\Delta$BA, point test (pp)              & 1.84 [1.23, 2.50]   & 2.24 [1.62, 2.90] \\
$\Delta$BA, block test (pp)              & 0.44 [$-0.11$, 1.03] & 1.10 [0.45, 1.81] \\
\quad $p$ (block test)                    & 0.20                & 0.013 \\
$\Delta$SE, both targets (pp)            & 6.4 [3.6, 9.4]      & 11.2 [7.3, 15.4] \\
Selection changed (\%)                    & 35.5                & 38.3 \\
$\Delta$BA at 5\% / 30\% (pp)            & $-0.03$ / 2.46      & $-0.03$ / 3.65 \\
Trend (pp per pp)                         & 0.091 [0.061, 0.124] & 0.145 [0.100, 0.192] \\
$\Delta$BA, ties excluded (pp; 62 datasets) & 1.23 [0.75, 1.78] & 1.93 [1.26, 2.69] \\
\bottomrule
\end{tabular}
\end{table}

\section{Discussion}
\label{sec:discussion}

\subsection{Principal Findings}

Under the controlled conditions of this study, validating under a missingness pattern different from the deployment pattern led to worse classifier selection. The average cost was modest, 1.14~pp of test balanced accuracy, but it was consistent in direction: 49 of the 64 datasets lost performance, and no dataset gained more than 1~pp. The cost was negligible at a 5\% missingness rate and increased approximately linearly with the rate, reaching 2.46~pp at 30\%. The same pattern held under non-wrapping linear blocks, where the cost was larger (1.67~pp overall) and reached significance for block-masked as well as point-masked tests. Missingness-pattern mismatch is therefore a source of selection error that adds to the error already present under matched validation, rather than a curiosity confined to a few datasets.

The size of the effect should be read against two reference points. First, matched validation was itself imperfect: in the $PP$ condition, the selected candidate was suboptimal in 26.9\% of conditions and lost 2.06~pp relative to the candidate-set oracle. Mismatch increased regret by roughly half of this matched-validation regret overall, and almost doubled it for point-masked tests ($BP$: 3.89~pp). Second, the cost of mismatch was small compared with the effect of the test pattern on achievable performance: the oracle was 11.1~pp lower on circular-block-masked than on point-masked tests. Mismatch matters for which candidate is chosen, but it does not by itself determine how difficult a deployment condition is.

\subsection{Asymmetry Between Target Patterns}

The cost of mismatch was concentrated in point-masked deployment. The candidate-level results suggest a mechanism, although the study was not designed to test it. On point-masked tests, MiniRocket-Ridge accounted for 84.0\% of the fractional oracle share, so departures from it were often costly. Block-masked validation moved selection away from MiniRocket-Ridge towards 1NN-DTW and the statistical-feature Random Forest, which were on average 10.5~pp and 14.5~pp below the oracle when selected in $BP$. On block-masked tests the oracle was more evenly distributed across candidates, and point validation's preference for MiniRocket-Ridge cost little, because MiniRocket-Ridge remained close to the best candidate on block-masked tests. The asymmetry therefore appears to depend on how concentrated the best candidate is under each deployment pattern, rather than on an intrinsic property of point or block missingness.

A related consideration may explain why the block-test effect reached significance only under linear blocks. Linear blocks never wrap, so interior positions are masked more often than boundary positions, unlike under point or circular block masking. Linear-block validation and deployment may therefore differ more from their point-masked counterparts, and the larger block-test cost under linear blocks (1.10~pp compared with 0.44~pp) is consistent with the hypothesis that the cost of mismatch grows with the dissimilarity between the validation and deployment conditions. Because the two constructions were analysed separately and not tested against each other, this comparison remains descriptive.

\subsection{Selection Change Versus Performance Loss}

Changing the validation pattern altered the selected candidate in about a third of target pairs, but a different selection was not necessarily a worse one: among changed selections, the mismatched choice was worse in 58.3\%, better in 37.3\%, and equally accurate in 4.4\%. At the 5\% rate, the selection changed in about one pair in five without any detectable loss. Selection-change frequencies therefore overstate the practical consequences of mismatch, and studies of method selection under distribution shift should report performance-based outcomes, such as the paired test-performance difference or regret, rather than agreement between selected methods alone.

\subsection{Relation to Prior Work}

Previous studies have shown that changes in the missingness process between development and deployment can degrade predictive performance~\cite{zhou2023domain,rockenschaub2024robust} and that the best imputation or prediction pipeline can depend on the missingness scenario~\cite{poette2026benchmarking,li2026prediction}. The present results add a complementary, selection-level observation: even when the training data are complete, the candidate set is fixed, and the missingness rate is matched, the temporal pattern of missingness in the validation data affects which classifier is chosen and how well it performs at deployment. Benchmark studies have emphasized that evaluation protocols influence conclusions about relative classifier performance~\cite{bagnall2017,middlehurst2024bakeoff}; our results indicate that the missingness pattern of the validation data is one such protocol choice.

\subsection{Practical Implications}

Three practical implications follow, within the scope of the studied setting. First, when the temporal pattern of missingness at deployment is known, validation data should be masked with the same pattern type, particularly at moderate to high missingness rates; at a 5\% rate the choice made little difference. Second, when contiguous gaps are possible at deployment but not represented in validation, or vice versa, practitioners should consider the possibility of a loss in deployment performance relative to matched validation and, where feasible, compare candidate rankings under both patterns before committing to a method. Third, because substantial selection error remained even under matched validation, the effects of validation-set size and repeated validation on selection reliability warrant separate investigation.

\section{Limitations}
\label{sec:limitations}

The conclusions are conditional on the controlled design, and several limitations restrict their generality.

\paragraph{Candidate set and imputation.} Only three prespecified candidates were compared, and their hyperparameters were fixed rather than tuned on validation data. A larger or more homogeneous candidate set could produce larger or smaller selection effects; in particular, the asymmetry discussed in \cref{sec:discussion} depends on the dominance of one candidate on point-masked tests. All series were imputed by linear interpolation, and other imputation methods, or classifiers that handle missing values natively, may interact differently with the missingness pattern.

\paragraph{Missingness mechanisms.} Missingness was synthetic and completely at random with respect to series values and labels~\cite{little2019statistical}, and only two pattern types were compared at matched rates. Real deployments may involve missingness that depends on the signal or the class, mixtures of isolated points and gaps, multiple gaps per series, or rates that differ between validation and deployment. The circular block construction serves experimental control; the linear-block analysis shows that the conclusions hold under one alternative construction but does not cover other gap-length distributions.

\paragraph{Training data.} The fitting subset was always complete. This isolates the effect of the validation pattern on selection, but in many applications the training data are also incomplete, and models trained on masked data may respond differently to the deployment pattern.

\paragraph{Datasets.} The 64 datasets are univariate, equal-length, originally complete UCR2015 tasks. Datasets with large combinations of sample size and series length were excluded for computational reasons, so the sample over-represents small and medium-sized tasks, in which validation subsets are small. Several datasets belong to related task families (for example, the phalanx outline tasks), and dataset-level resampling does not model this dependence. Bootstrap intervals and tests therefore describe the included tasks, not a probability sample of all TSC problems.

\paragraph{Validation size and ties.} The stratified validation subsets were small for some datasets, which contributes to selection error under both matched and mismatched validation. Some splits omit a class from validation, and validation ties occurred in 16.7\% of primary conditions. The tie-excluded analysis supports the main result, but it conditions on tie-free observations rather than reproducing the original aggregation.

\paragraph{Oracle and inference.} The candidate-set oracle is the observed best of three candidates on a finite test set, not the population-optimal method, and selection error indicates observed rather than statistically established suboptimality. Multiplicity was controlled only across rates within each metric; overall, stratum-specific, trend, and supplementary analyses were not jointly corrected. Descriptive outputs from 27 and 46 datasets were inspected before the primary run was complete, although no inferential test was performed on them and the dataset list had been fixed in advance.

\paragraph{Implementation.} The primary and supplementary runs used different execution paths. Their equivalence was verified at the level of recorded point-side metrics and by regression tests (\cref{sec:implementation}), not by comparing every prediction array or fitted estimator.

\section{Conclusion}
\label{sec:conclusion}

This study examined whether a mismatch between validation and test missingness patterns affects validation-based classifier selection in time-series classification. In a controlled $2 \times 2$ design over 64 UCR datasets, with complete training data, a fixed candidate set, and linear interpolation, selecting a classifier under a mismatched pattern reduced the test balanced accuracy of the selected candidate by 1.14~pp on average. The effect was positive for most datasets, negligible at a 5\% missingness rate, and increased with the rate to 2.46~pp at 30\%. It was concentrated in point-masked deployment, where block-masked validation shifted selection away from the candidate that was usually best. A supplementary analysis with non-wrapping linear blocks reproduced these findings with a larger effect, including for block-masked deployment.

Mismatched validation frequently changed the selected classifier, but a changed selection did not always reduce performance, so the consequences of mismatch are better measured by paired test performance and regret than by selection agreement. When the expected deployment missingness pattern is known, matching the validation pattern to it provides a simple safeguard, particularly at higher missingness rates. Future work could investigate robust method selection when deployment missingness is uncertain or changes over time, as well as extend the design to tuned and larger candidate sets, alternative imputation methods, incomplete training data, value-dependent missingness, and multivariate series.




\section*{Data Availability}

All datasets analysed in this study are publicly available from the UCR Time Series Classification Archive~\cite{dau2019ucr} and were accessed through aeon~1.6.0. The classifier-level and selection-level result files, analysis tables, and verification records underlying the reported results are retained with the project repository described under Code Availability.


\section*{Code Availability}

The code used to generate masks, run the experiments, and analyse the results, together with the experimental configurations, is maintained in the repository \url{ https://github.com/neko-nikki-research/tsc-missingness-mismatch} and will be made publicly available upon publication.


\section*{Conflict of Interest}

The authors declare no conflict of interest.


\printbibliography

\end{document}